\documentclass[10pt,twocolumn,letterpaper]{article}

\usepackage{cvpr}
\usepackage{times}
\usepackage{epsfig}
\usepackage{graphicx}
\usepackage{amsmath}
\usepackage{amssymb}

\usepackage{subcaption}
\usepackage[titletoc]{appendix}

\usepackage[breaklinks=true,bookmarks=false,colorlinks=true]{hyperref}

\cvprfinalcopy % *** Uncomment this line for the final submission

\def\cvprPaperID{****} % *** Enter the CVPR Paper ID here

\ifcvprfinal\fi
\begin{document}

%%%%%%%%% TITLE
\title{Measuring the tendency of CNNs to Learn Surface Statistical Regularities}

\author{Jason Jo\\
MILA, Universit\'{e} de Montr\'{e}al\\
IVADO \\
{\tt\small jason.jo.research@gmail.com}
\and
Yoshua Bengio\\
MILA, Universit\'{e} de Montr\'{e}al \\
CIFAR\\
{\tt\small yoshua.umontreal@gmail.com}
}

\maketitle

%%%%%%%%% ABSTRACT
\begin{abstract}
Deep CNNs are known to exhibit the following peculiarity: on the one hand they generalize extremely well to a test set, while on 
the other hand they are extremely sensitive to so-called adversarial perturbations. The extreme sensitivity of high performance CNNs to 
adversarial examples casts serious doubt that these networks are learning high level abstractions in the dataset. We are concerned with the 
following question: How can a deep CNN that does not learn any high level semantics of the dataset manage to generalize so well? The goal 
of this article is to measure the tendency of CNNs to learn surface statistical regularities of the dataset. To this end, we use Fourier 
filtering to construct datasets which share the exact same high level abstractions but exhibit qualitatively different surface statistical 
regularities. For the SVHN and CIFAR-10 datasets, we present two Fourier filtered variants: a low frequency variant and a randomly filtered 
variant. Each of the Fourier filtering schemes is tuned to preserve the recognizability of the objects. Our main finding is that CNNs 
exhibit a tendency to latch onto the Fourier image statistics of the training dataset, sometimes exhibiting up to a 28\% generalization gap 
across the various test sets. Moreover, we observe that significantly increasing the depth of a network has a very marginal impact on 
closing the aforementioned generalization gap. Thus we provide quantitative evidence supporting the hypothesis that deep CNNs tend to learn 
surface statistical regularities in the dataset rather than higher-level abstract concepts.
\end{abstract}

\section{Introduction} 

The generalization ability of a machine learning model can be measured by evaluating its accuracy on a withheld test set. For visual 
learning tasks, convolutional neural networks (CNNs) \cite{OriginalCNN} have become the de facto machine learning model. These CNNs have 
achieved record breaking object recognition performance for the CIFAR-10 \cite{cifar-10}, SVHN \cite{SVHN} and ImageNet 
\cite{ImageNet} datasets, at times surpassing human performance \cite{DelvingDeeper}. Therefore, on the one hand, very deep CNN 
architectures have been designed which obtain very good generalization performance. On the other hand, it has been shown that these same 
CNNs exhibit an extreme sensitivity to so-called \emph{adversarial examples} \cite{adv-examples}. These adversarial examples are 
perceptually quite similar to the original, ``clean'' image. Indeed humans are able to correctly classify the adversarial image with 
relative ease, whereas the CNNs predict the wrong label, usually with very high confidence. The sensitivity of high performance CNNs to 
adversarial examples casts serious doubt that these networks are actually learning high level abstract concepts \cite{FawziAnalysis, 
EasilyFooled}. This begs the following question: \emph{How can a network that is not learning high level abstract concepts manage to 
generalize so well?}

Roughly speaking, there are two ways in which a machine learning model can generalize well. The first way is the ideal way: the 
model is trained in a manner that captures high level abstractions in the dataset. The second way is less than 
ideal: the model has a tendency to overfit to superficial cues that are actually present in both the train and test datasets; thus the 
statistical properties of the dataset plays a key role. In this fashion, high performance generalization is possible without 
actually explicitly learning any high level concepts.

In Section 2 we discuss the generalization ability of a machine learning model and its relation to the surface 
statistical regularities of the dataset. In particular, by drawing on computer vision literature on the statistics of natural images, we 
believe that it is possible for natural image train and test datasets to share \emph{many superficial cues}. This leads us to postulate our 
main hypothesis: \textbf{\emph{the current incarnation of deep neural networks has a tendency to learn surface statistical regularities in 
the dataset}}. In Section 3 we discuss related work.

To test our hypothesis, we will quantitatively measure this tendency. To this end, for a dataset $X$ it is sufficient to construct a 
perturbation map $F$:
\begin{equation}\label{eg:perturbation-map}
F : X \mapsto X',
\end{equation}
which satisfies the following properties:
\begin{enumerate}
 \item \emph{Object Recognizability is preserved}. Given a clean image $x \in X$ and its perturbation $x' \in X'$, the 
recognizability of the object in the images $x$ and $x'$ is almost exactly preserved from the perspective of a human. This guarantees that 
$X,X'$ share the same high level concepts. 
 \item \emph{Qualitatively Different Surface Regularities}. While the recognizability of the objects is roughly preserved by the 
perturbation map $F$, the datasets $X$ and $X'$ also exhibit qualitatively different image statistics. In combination with the first 
property, this guarantees that the two datasets $X,X'$ share the same high level abstractions but may exhibit different superficial cues. 
 \item \emph{Existence of a non-trivial generalization gap}. Given the clean dataset $\{(X_\mathrm{train}, Y_\mathrm{train}), 
(X_\mathrm{test}, Y_\mathrm{test})\}$, the map $F$ produces another dataset $\{(X'_\mathrm{train}, Y_\mathrm{train}), 
(X'_\mathrm{test}, Y_\mathrm{test})\}$. Now we simply measure the test accuracy of a deep CNN trained on either $X_\mathrm{train}$ or 
$X'_\mathrm{train}$ on both $X_\mathrm{test}$ and $X'_\mathrm{test}$ and compute the corresponding generalization gap. A good 
perturbation map $F$ is one in which the generalization gap is non-trivial.
\end{enumerate}

In Section 4, we show that a natural candidate for these maps $F$ are maps which are defined by Fourier filtering. We define two types of 
Fourier filtering schemes: radial and random. Each of these schemes has a parameter that needs to be tuned: for the radial filter we must 
define the radius of the mask, and for the random filter we must define the probability of setting a Fourier mode to zero in a uniformly 
random fashion. We will present our tuned Fourier filter maps for the SVHN and CIFAR-10 datasets. In addition we present visual evidence 
that the recognizability of the objects in the filtered datasets is extremely robust to the human eye. Due to the fact that we are Fourier 
filtering, the filtered datasets will by construction exhibit different image statistics. Thus we are able to produce filtered/perturbed 
training and test datasets which share the same high level perceptual content as the original datasets but exhibit qualitatively different 
surface statistical regularities, \eg the Fourier image statistics. 

In Section 5 we present generalization experiments which are designed to test our main hypothesis. 
High performance CNNs are trained on one of the $\{X_{\mathrm{unfiltered}}, X_{\mathrm{radial}}, X_\mathrm{random}\}$ datasets and 
the test accuracy is evaluated on \emph{all} the other test distributions. We show a general pattern of the deep CNN models exhibiting a 
tendency 
to latch onto the surface statistical regularities of the training dataset, sometimes exhibiting up to a \emph{28\% gap in test 
accuracy}. Another striking result was that CNNs trained on $X_\mathrm{unfiltered}$ generalized quite poorly to $X_\mathrm{radial}$, to 
which we report a generalization gap upwards of 18\%. Moreover, increasing the depth of the CNN in a significant manner (going from 92 
layers to 200 layers) has a very small effect on closing the generalization gap. 

Our last set of experiments involves training on the fully augmented training set, which now enjoys a variance of its Fourier image 
statistics. We note that this sort of data augmentation was able to close the generalization gap. However, we stress that it is doubtful 
that this sort of data augmentation scheme is sufficient to enable a machine learning model to truly learn the semantic concepts present in 
a dataset. Rather this sort of data augmentation scheme is analogous to adversarial training \cite{adv-examples, HarnessingAdvExamples}: 
there is a non-trivial regularization benefit, but it is not a solution to the underlying problem of not learning high level semantic 
concepts, nor do we aim to present it as such.

In Section 6 we present our conclusion that our empirical results provide evidence for the claim that the current incarnation of 
deep neural networks are not actually learning high level abstractions. Finally we highlight promising new research directions towards this 
goal.

\section{Generalization and Surface Statistical Regularities}

In this section we wish to reconcile two seemingly inharmonious yet individually valid (in their respective contexts) claims about the 
generalization properties of deep CNNs:
\begin{enumerate}
 \item \emph{Claim \#1:} Deep CNNs are generalizing extremely well to an unseen test set. 
 \item \emph{Claim \#2:} General sensitivity to adversarial examples show that deep CNNs are not truly capturing abstractions in the 
dataset. 
\end{enumerate}

One key intuition to understand the above two claims is to recognize that \emph{there is actually a strong statistical relationship between 
image statistics and visual understanding}. For example \cite{Torralba:2003} explored the relationship between image statistics and visual 
scene categorization. They were actually able to use image statistics to predict the presence or absence of objects in an image of a 
natural scene. In \cite{Pinto}, the authors placed synthetic, rendered objects into statistically unlikely to occur yet natural looking 
backgrounds. For example Figure 2 of \cite{Pinto} depicts a car floating at an angle in a grassy area with clouds in the background. They 
then used these synthetic images to test the invariance of various visual features for object recognition, one example being the 
sensitivity of object recognition to covariation with the background. To this end, they hypothesize that computer vision algorithms 
may ``\emph{...lean heavily on background features to perform categorization.}'' 

Therefore, when the training and test set share similar image statistics, it is wholly 
possibly for a machine learning model to learn \emph{superficial cues and generalize well, albeit in a very narrow sense as they are 
highly dependent on the image statistics}. Adversarial examples would be destroying the superficial cues. We believe that this is precisely 
how deep CNNs can attain record breaking generalization performance on all sorts of natural image tasks, and yet can be so sensitive to 
adversarial perturbations. Most importantly, the above reasoning can explain how a machine learning model can actually generalize well 
without ever having to explicitly learn abstract concepts. To this end, we formally state our main hypothesis: 
\begin{quote} \textbf{\emph{The current incarnation of deep neural networks exhibit a tendency to learn 
surface statistical regularities as opposed to higher level abstractions in the dataset. For tasks such as object recognition, due to the 
strong statistical properties of natural images, these superficial cues that the deep neural network have learned are sufficient for high 
performance generalization, but in a narrow distributional sense.}}
\end{quote}
Having stated our main hypothesis, we feel the need to stress that it is not fair to compare the generalization performance of CNN to a 
human being. In contrast to a CNN, a human being is exposed to an incredibly diverse range of lighting conditions, viewpoint variations, 
occlusions, among a 
myriad of other factors.

\section{Related Work} 

To the best of our knowledge, we are the first to consider using Fourier filtering for the purpose of measuring the tendency of 
CNNs to learn surface statistical regularities in the dataset. With respect to related work, we highlight \cite{NegativeMNIST} which showed 
that CNNs trained on the clean MNIST, CIFAR-10 and GTSRB \cite{GTSRB} datasets generalize quite poorly to the so-called ``negative'' test 
sets, where the test images have negated brightness intensities. The major difference from \cite{NegativeMNIST} and our work is that 
negative images are known to be more challenging for the task of object recognition for human beings, we refer to \cite{NegativeImages, 
NegativeFaces, GalperNegative, LowDataRates} and the numerous references therein. Indeed from \cite{NegativeImages} we quote: 
``\emph{...negative  images containing low-frequency components 
were considerably more difficult to recognize than the corresponding positive images.}'' From \cite{Field:87, Burton:87}, we know that 
natural images tend to have the bulk of their Fourier spectrum concentrated on the low to mid range frequencies. To this end, we view our 
Fourier filtering scheme as a better principled scheme than image negation with respect to preserving the recognizability of the objects. 
Finally \cite{NegativeMNIST} employs CNN models for the CIFAR-10 which attain a maximum test accuracy of about 84\% while we use a more 
modern and up to date CNN model, regularly achieving 95\% test accuracy for the CIFAR-10, much closer to the current state of the art.

The problems of transfer learning \cite{YosinskiTransfer, TransferSurvey, CNNTransfer} and domain adaptation \cite{DomainAdapt, 
DomainAdapt2} both investigate the generalization ability of deep neural networks. However we believe the more relevant research literature 
is coming from adversarial examples. Originally introduced and analyzed for the MNIST dataset in \cite{adv-examples}, adversarial examples 
have sparked a flurry of research activity. The original work \cite{adv-examples} showed that it is not only possible to generate 
adversarial examples which fool a given network, these adversarial examples actually \emph{transfer} across different network 
architectures. \cite{HarnessingAdvExamples} further explored the transferability of adversarial examples for the MNIST and CIFAR-10 
datasets, and \cite{UniversalAdv} was able to show the existence of a universal adversarial noise for ImageNet dataset. The universal 
adversarial noise is image agnostic, \eg is able to be applied to a wide range of images and still fool various networks. 

As a response to these adversarial examples, there have been various attempts to increase the robustness of deep neural networks to 
adversarial perturbations. \cite{PapernotDistillation} employed defensive-distillation. \cite{ContractiveDefense} used so-called
contractive networks. Moreover, \cite{ContractiveDefense} posited that the core problem with adversarial examples emanates from the 
current training setup used in deep learning, rather than the network architecture.  Along these lines, \cite{Parseval} obtained 
promising results by modifying the training regime of SGD by forcing the convolutional layers to be approximate Parseval tight frames 
\cite{Kovacevic}. This method led to state of the art performance on the CIFAR-10/100 as well as the SVHN dataset while also increasing the 
robustness of the network to adversarial examples. Similarly in \cite{BANG} the training loop is modified to improve robustness to 
adversarial examples. Specifically \cite{BANG} modifies SGD by rescaling the batch gradients and reports an increased robustness to 
adversarial examples for the MNIST and CIFAR-10 datasets, though the CIFAR-10 models suffer a non-trivial degradation in clean test 
accuracy.

The most popular adversarial robustness method has been adversarial training \cite{HarnessingAdvExamples, DistSmoothing, DeepFool, 
ShahamAdv, adv-examples, KurakinAdv}. Adversarial training methods all rely on data augmentation: the training data set is augmented with 
adversarial examples. In general we comment that these methods tend to rely on a certain adversarial example generation technique. Thus 
these methods are not guaranteed to be robust to adversarial examples generated from some alternate method. To this end, 
\cite{FawziAnalysis, EasilyFooled} cast doubt that supervised CNNs are actually learning semantic concepts in the datasets.

\section{Robustness of Object Recognition to Fourier Filtering}

The surface statistical regularities we will be 
concerned with are the Fourier image statistics of a dataset. While natural images are known to exhibit a huge variance in the raw pixel 
space, it has been shown \cite{Field:87, Burton:87} that the Fourier image statistics of natural images obey a power law decay: the power 
$P(w)$ of a Fourier mode (also referred to as a frequency) $w$ decays $\propto \frac{A}{|w|^{2-\eta}}$ for some $A$ and $\eta$ which varies 
over the image types, but $\eta$ is typically small. An immediate takeaway is that natural images tend to have the bulk of their Fourier 
spectrum concentrated in the low to medium range frequencies. \iffalse See Figure \ref{fig:cifar10_log_pow_spec} for a visual 
representation 
of the Fourier power spectrum for some CIFAR-10 images. \fi \emph{Due to this power law concentration of energy in the frequency space, it 
is 
possible to perform certain types of Fourier filtering and preserve much of the perceptual content of an image.}

\iffalse
\begin{figure}[ht!]
  \begin{center}
    \includegraphics[width=0.99\linewidth]{./cifar10_log_pow_spec.png}
  \end{center}
  \caption{\textbf{Top row}: CIFAR-10 images. \textbf{Bottom row}: Log power spectrum of the red channel for each of the corresponding 
	  images. The 2D Fast Fourier Transform has been shifted so that the DC component is the center of the image. Best viewed in color.}
  \label{fig:cifar10_log_pow_spec}
\end{figure} 
\fi

In this section we define our Fourier filtering setup and present visual evidence that while the Fourier filtering process does indeed 
introduce artifacts and does degrade the quality of the image, the recognizability of the objects in the image is still quite robust from a 
human point of view. For a given dataset $(X,Y)$ 
(which can represent either the train or test dataset), we have the following:
\begin{itemize}
 \item $(X,Y)$ itself, the \emph{unfiltered} version. 
 \item $(X_\mathrm{radial}, Y)$, the \emph{low frequency filtered} version. We use a radial mask in the Fourier domain to set higher 
frequency modes to zero. 
 \item $(X_\mathrm{random}, Y)$, the \emph{randomly filtered} version. We use a random mask which uniformly at random sets a Fourier 
mode to zero with probability $p$. The random mask is generated once and applied to all the elements of $X$. 
\end{itemize}

Let $X \in \mathbb{R}^{H \times W}$ denote a $H \times W$ shaped image with only 1 color channel. Recall that the 2D Discrete Fourier 
Transform (DFT) \cite{FFT-book} of $X$, denoted $\mathcal{F}(X)$ is defined as: 
\begin{equation}
 \mathcal{F}(X)[k,l] := \frac{1}{\sqrt{HW}}\sum_{h=0}^{H-1} \sum_{w=0}^{W-1} X[w,h] e^{-j 2\pi( \frac{wk}{W} + \frac{lh}{H} )},
\end{equation}
for $k = 0, \dots, W-1, l = 0, \dots, H-1$, and $j = \sqrt{-1}$. If $X$ is an RGB image, so it has $C=3$ channels, we then compute the DFT 
for each image channel. We will furthermore consider the \emph{shifted DFT} in which the DC component is located in the center of the 
$H\times W$ matrix as opposed to the $(0,0)$ index. We let $\mathcal{F}^{-1}(X)$ denote the inverse FFT of $X$, composed with the 
appropriate inverse spectral shift. 

In this article, we will consider two types of Fourier filtering schemes:
\begin{itemize}
 \item \emph{Radial masking}. This scheme is parameterized by the mask radius $r$. We will require that each of our images $X$ 
have height $H$ and width $W$ of even length. The radial mask $M_r$ is defined as:
\begin{equation}\label{eg:radial_mask}
 M_r[i,j] := \begin{cases}
             1 & \textrm{ if } \| (i,j) - (W/2,H/2)\|_{\ell_2} \leq r, 
              \\ 0 & \textrm{ otherwise. }  
             \end{cases}
\end{equation}
We use $\|x-y\|_{\ell_2}$ to denote the $\ell_2$ distance between the vectors $x$ and $y$. The mask $M_r$ is applied across the channels. 

For an unfiltered dataset $X$, we define $X_\mathrm{radial}$ as:
\begin{equation}\label{eq:x_low_def}
 X_{\mathrm{radial}} := \mathcal{F}^{-1}( \mathcal{F}(X)\circ M_r),
\end{equation}
where $\circ$ denotes the element-wise \emph{Hadamard product}.

 \item \emph{Uniformly random masking}. This scheme is parameterized by a drop probability $p$. We will generate a random mask $M_p$ once, 
and then apply the same random mask to each of the DFTs. The mask $M_p$ is defined as:
\begin{equation}\label{eq:random_mask}
M_p[c,i,j] := \begin{cases}
             0 & \textrm{ with probability } p,
             \\ 1 & \textrm{ otherwise.}
            \end{cases}
\end{equation}
Note that we do not have the same random mask per channel. For an unfiltered dataset $X$, we define $X_\mathrm{random}$ as:
\begin{equation}\label{eq:x_random_def}
 X_{\mathrm{random}} := \mathcal{F}^{-1}( \mathcal{F}(X)\circ M_p),
\end{equation}
\end{itemize}

For the rest of the section we will present which mask radius and random masking probability hyperparameters were used for the SVHN and 
CIFAR-10 natural image datasets. Note that we did not use the MNIST dataset due to its extreme sparsity which results in very low 
recognizability of the digits after Fourier filtering. 

\subsection{SVHN}
 \begin{figure}[ht!]
 \begin{center}
\includegraphics[scale=.41]{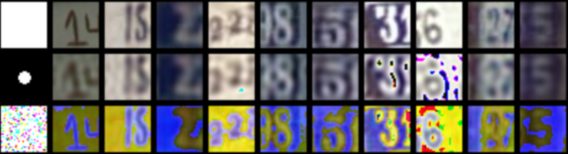}
\end{center}
   \caption{The first image in each column corresponds to the Fourier mask in frequency space. A white pixel corresponds to preserving the 
Fourier 
mode, black/color corresponds to setting it to zero. \textbf{Top row}: No Fourier filtering applied, original 
SVHN images. \textbf{Middle row}: Radial mask and the corresponding filtered images. \textbf{Bottom row}: Random mask and the corresponding 
filtered images. Best viewed in color.}
\label{fig:svhn_fourier_masks}
\end{figure}
For the SVHN dataset, the images have spatial shape $(32,32)$ with 3 color channels corresponding to RGB. For the radial masking we used a 
mask radius of 4.25 and for random masking we used $p=0.1$. In Figure \ref{fig:svhn_fourier_masks} we show the masks and a comparison 
between unfiltered and filtered images. We notice that while both the radial and random filters produce some visual artifacts, and some 
random masks can actually result in noticeable color distortions, the overall recognizability of the digits is quite robust.

\subsection{CIFAR-10}
 \begin{figure}[ht!]
 \begin{center}
\includegraphics[scale=.42]{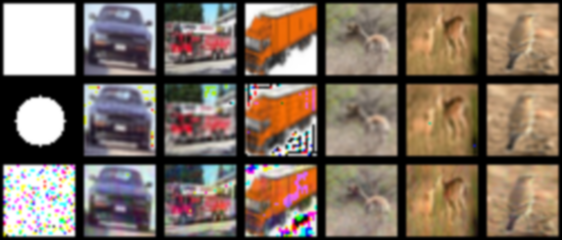}
\end{center}
   \caption{The first image in each column corresponds to the Fourier mask in frequency space. A white pixel corresponds to preserving the 
Fourier 
mode, black/color corresponds to setting it to zero. \textbf{Top row}: No Fourier filtering applied, original 
CIFAR-10 images. \textbf{Middle row}: Radial mask and the corresponding filtered images. \textbf{Bottom row}: Random mask and the 
corresponding filtered images. Best viewed in color.}
\label{fig:cifar_10_fourier_masks}
\end{figure}

For the CIFAR-10, the images have spatial shape $(32,32)$ with 3 color channels corresponding to RGB. For the radial masking we used a mask 
radius of 11.0 and for random masking we used $p=0.1$. In Figure \ref{fig:cifar_10_fourier_masks} we show the masks and a comparison 
between 
unfiltered and filtered images. Observe that while there are undoubtedly artifacts that arise from our Fourier filters, the recognizability 
of the objects is quite robust to the human eye. Furthermore, the artifacts that do occur have a tendency to actually occur \emph{in the 
background of the image} or cause minimal degradation to the recognizability of the object. Refer to the Supplementary Materials for more 
visual examples.

\section{Generalization Experiments}

In this section we present our generalization experiments. Our generalization experiments consists of the following: for the CIFAR-10 and 
SVHN datasets, we take some established high-performance CNN architectures (while they are not state of the art, they are typically very 
close to state of the art performance) and train them on one of the following: $\{X_{\mathrm{train}}^{\mathrm{unfiltered}}, 
X_{\mathrm{train}}^{\mathrm{radial}}, X_{\mathrm{train}}^{\mathrm{random}}\}$ and then test the accuracy of each of these trained models on 
\emph{all} of the following test sets $\{X_{\mathrm{test}}^{\mathrm{unfiltered}}, X_{\mathrm{test}}^{\mathrm{radial}}, 
X_{\mathrm{test}}^{\mathrm{random}}\}$. We refer to a \emph{test gap} or a \emph{generalization gap} as the maximum difference of the 
accuracy on the various test sets. For the CIFAR-10 and SVHN datasets, we use exactly the parameterized Fourier 
filtering setups from Section 4.

For both the SVHN and CIFAR-10 experiments we trained a Preact ResNet \cite{Preact} with Bottleneck architecture of depth 92 and 200 using 
the so-called ``HeNormal'' initialization from \cite{resnet}, using a random seed of 0. For formatting purposes, we only show graphical 
plots for the very deepest highest performance (Preact-ResNet-200) model and merely summarize the Preact-ResNet-92 models performance in a 
table. We include the full graphical plots for the depth 92 model in the Supplementary Materials.

In general, none of the training sets generalized universally well to the various test sets. So we also trained on the fully augmented 
training set:
\begin{equation}\label{eqaution:augmented}
X_{\mathrm{train}}^{\mathrm{augmented}} :=
X_{\mathrm{train}}^{\mathrm{unfiltered}} \cup 
X_{\mathrm{train}}^{\mathrm{radial}} \cup X_{\mathrm{train}}^{\mathrm{random}}
\end{equation}
and then measured the generalization performance on the various test sets. 

\subsection{SVHN Experiments}

For the SVHN dataset we follow the convention of \cite{WRN}: we combine the original training set and the extra dataset to form a new 
training set and we normalize all the pixel intensities to [0,1] by dividing all the values by 255.0. 
Otherwise, no other form of data augmentation or pre-processing was used. 

We train the ResNets for 40 epochs using Nesterov momentum \cite{nesterov} with an initial learning rate of 0.01 with momentum parameter 
0.9. The training batchsize was 128, the L2 regularization parameter was 0.0005 and we decayed the learning rate at epochs 20 and 30 by 
dividing by 10. 

\begin{figure}
  \centering
  \begin{subfigure}[b]{0.95\linewidth}
    \centering
    \includegraphics[width=\linewidth]{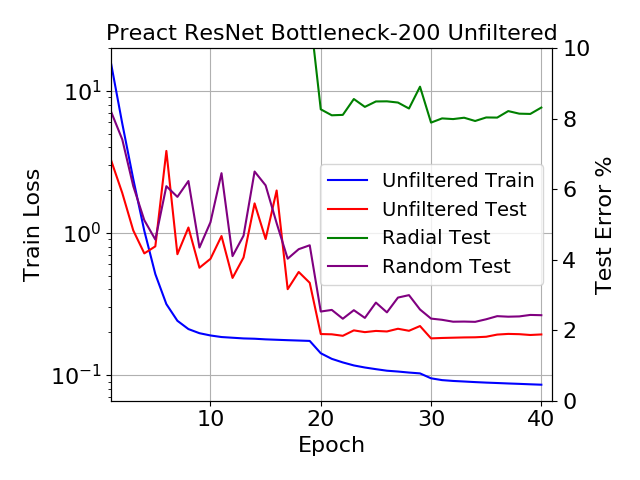}
    \caption[]%
    {{\small Trained on Unfiltered SVHN}}    
    \label{fig:resnet-200-svhn-10-unfiltered}
  \end{subfigure}
  \hfill
  \begin{subfigure}[b]{0.95\linewidth}
    \centering
    \includegraphics[width=\linewidth]{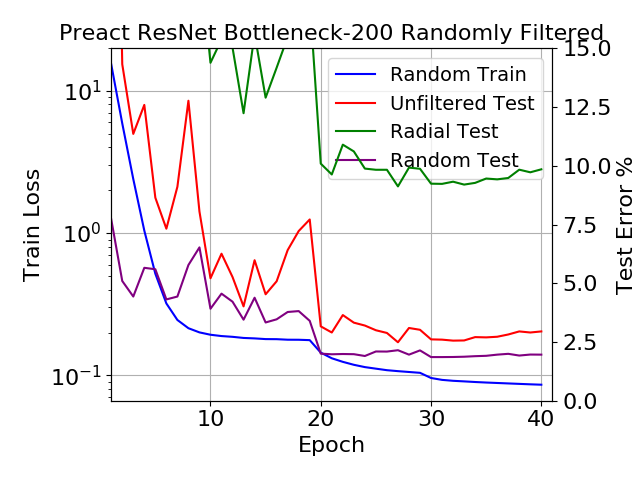}
    \caption[]%
    {{\small Trained on Randomly Filtered SVHN}}    
    \label{fig:resnet-200-svhn-random}
  \end{subfigure}
  \hfill
  \begin{subfigure}[b]{0.95\linewidth}   
    \centering 
    \includegraphics[width=\linewidth]{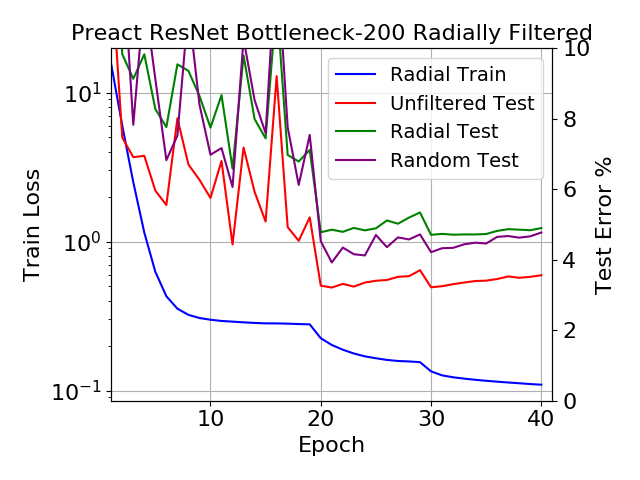}
    \caption[]%
    {{\small Trained on Radially Filtered SVHN}}    
    \label{fig:resnet-200-svhn-radial}
  \end{subfigure}
  \caption[]%
  {\small Generalization plots for Preact-ResNet-Bottleneck-200 model. \textbf{(a)} Trained on unfiltered SVHN. \textbf{(b)} 
  Trained on randomly filtered SVHN data. \textbf{(c)} Trained on radially filtered SVHN data. Best viewed in color.} 
  \label{fig:resnet-200-svhn}
\end{figure}

\begin{figure}
  \centering
  \includegraphics[width=\linewidth]{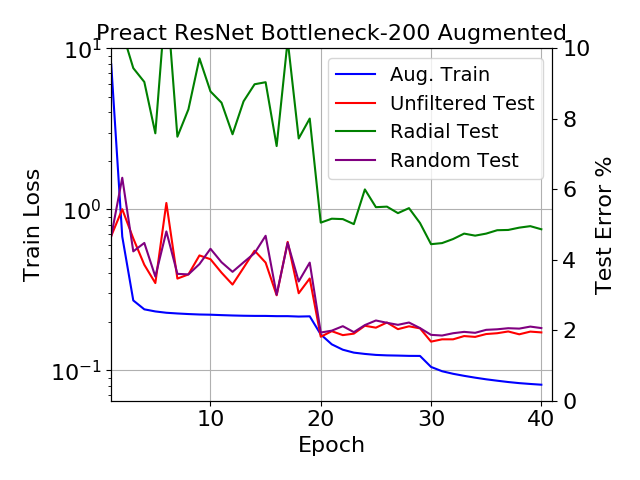}
  \caption[]%
  {\small Augmented SVHN Training plots for ResNet-200. Best viewed in color.} 
  \label{fig:resnet-200-svhn-augmented}
\end{figure}

In Figure \ref{fig:resnet-200-svhn} we present the generalization plots for the Preact-ResNet-200 trained on the unfiltered, randomly 
filtered and radially filtered training sets. In Figure \ref{fig:resnet-200-svhn-augmented} we present the graphical plot of the 
generalization curves for the Preact-ResNet-200 trained on the fully augmented training set. Finally in Figure 
\ref{fig:svhn-generalization-table} we present the SVHN generalization error table for both the Preact-ResNet-92 and the Preact-ResNet-200 
model.

From these figures we observe that when trained on the unfiltered data and tested on the radially filtered test data, the networks 
exhibited a generalization gap of about 6.4\%. Furthermore, when the nets were trained on the randomly filtered data, these 
networks had the worst generalization gap at approximately 7.89\%, again for the radially filtered test set. Training on the 
radially filtered dataset seems to enjoy a regularization benefit with respect to the unfiltered test set, actually generalizing nearly 
1.5\% better on the unfiltered test set than the radially filtered test set. We observe that the networks trained on the radially filtered 
data tend to have the lowest generalization gap, and furthermore that training on the augmented training set reduced the generalization 
gap. One general theme was that regardless of the training set, depth seemed to have a negligible effect on closing the generalization gap.
    
\begin{figure}
  \captionsetup[subfigure]{justification=centering}
  \centering
  \begin{subfigure}[b]{0.99\linewidth}
    \centering
    % stuff
    \begin{tabular}{|c|c|c|c|c|}
      \hline
      Train/Test & Unfilt. & Radial & Random & Gen. Gap\\ \hline
      Unfilt. & 1.95\%  & 8.41\% & 2.68\% & 6.46\%\\ \hline
      Radial & 3.50\% & 5.07\% & 5.67\% & 2.17\% \\ \hline
      Random & 4.01\% & 11.90\% & 2.04\% & 7.89\%\\ \hline
      Augmented & 2.11\% & 5.06\% & 2.15\% & 2.95\% \\
      \hline
    \end{tabular}
    \caption[]%
    {{\small Preact-ResNet-Bottleneck-92 SVHN Generalization}}    
    \label{fig:resnet-92-svhn-table}
  \end{subfigure}
  \hfill
  \begin{subfigure}[b]{0.95\linewidth}
    \centering
    % stuff
    \begin{tabular}{|c|c|c|c|c|}
      \hline
      Train/Test & Unfilt. & Radial & Random & Gen. Gap\\ \hline
      Unfilt. & 1.88\% & 8.31\% & 2.42\% & 6.43\%\\ \hline
      Radial & 3.56\% & 4.90\% & 4.77\% & 1.34\%\\ \hline
      Random & 2.95\% & 9.85\% & 1.96\% & 7.89\%\\ \hline
      Augmented & 1.94\% & 4.87\% & 2.06\% & 2.93\% \\
      \hline
    \end{tabular}
    \caption[]%
    {{\small Preact-ResNet-Bottleneck-200 SVHN Generalization}}    
    \label{fig:resnet-200-svhn-table}
  \end{subfigure}
  \caption[]%
  {\small SVHN Generalization Table. The train indicates what the training set was, the test indicates the test set. Final test 
    error is listed and the corresponding generalization gap.} 
\label{fig:svhn-generalization-table}
\end{figure}

\subsection{CIFAR-10 Experiments} 

For the CIFAR-10 we perform global contrast normalization (zero-centering the training set and dividing by 
the pixel-wise standard deviation) and augment via horizontal flips. During training we pad each 32x32 image with zeros to a 40x40 image 
and extract a random 32x32 crop.

We train the ResNets for 100 epochs with an initial learning rate of 0.01, which we boost up to 0.1 after 400 updates. The momentum 
parameter is 0.9. The training batchsize was 128, the L2 regularization parameter was 0.0001 and we decayed the learning rate at epochs 50 
and 75 by dividing by 10. For the augmented models we trained a bit longer: 120 epochs with learning rate decays at epochs 60 and 80.

\begin{figure}
  \centering
  \begin{subfigure}[b]{0.95\linewidth}
    \centering
    \includegraphics[width=\linewidth]{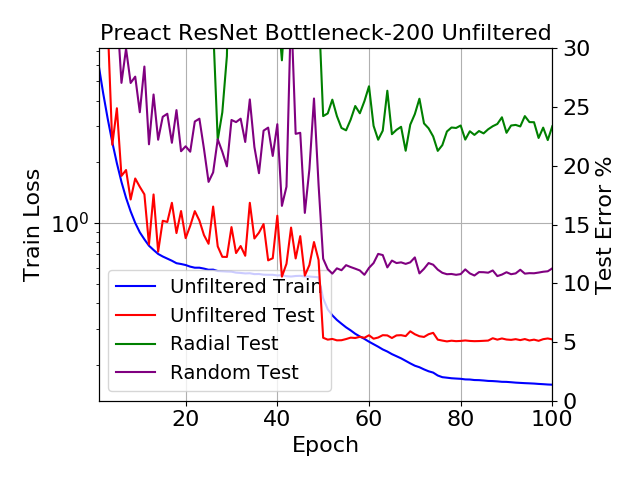}
    \caption[]%
    {{\small Trained on Unfiltered CIFAR-10}}    
    \label{fig:resnet-200-cifar-10-unfiltered}
  \end{subfigure}
  \hfill
  \begin{subfigure}[b]{0.95\linewidth}
    \centering
    \includegraphics[width=\linewidth]{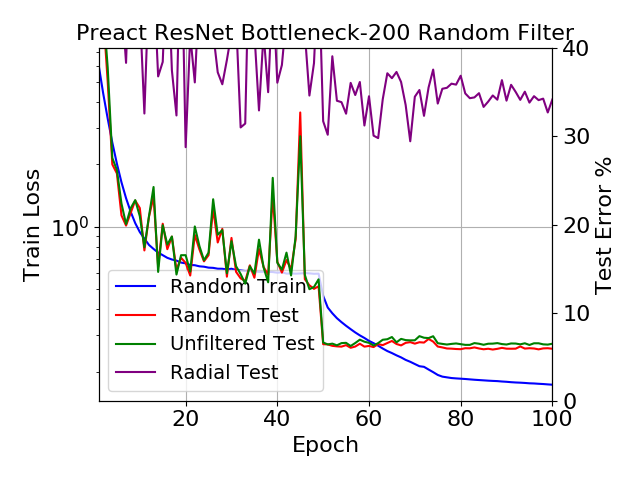}
    \caption[]%
    {{\small Trained on Randomly Filtered CIFAR-10}}    
    \label{fig:resnet-200-cifar-10-random}
  \end{subfigure}
  \hfill
  \begin{subfigure}[b]{0.95\linewidth}   
    \centering 
    \includegraphics[width=\linewidth]{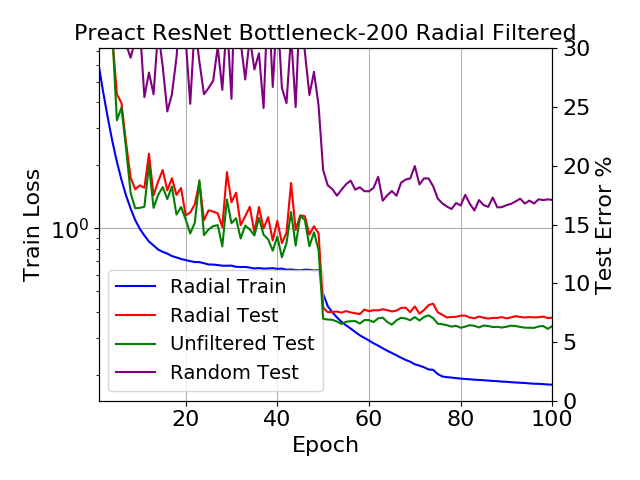}
    \caption[]%
    {{\small Trained on Radially Filtered CIFAR-10}}    
    \label{fig:resnet-200-cifar-10-radial}
  \end{subfigure}
  \caption[]%
  {\small Generalization plots for Preact-ResNet-Bottleneck-200 model. \textbf{(a)} Trained on unfiltered CIFAR-10 data. \textbf{(b)} 
  Trained on randomly filtered CIFAR-10 data. \textbf{(c)} Trained on radially filtered CIFAR-10 data. Best viewed in color.} 
\label{fig:resnet-200-cifar-10}
\end{figure}
    
\begin{figure}
  \centering
  \includegraphics[width=\linewidth]{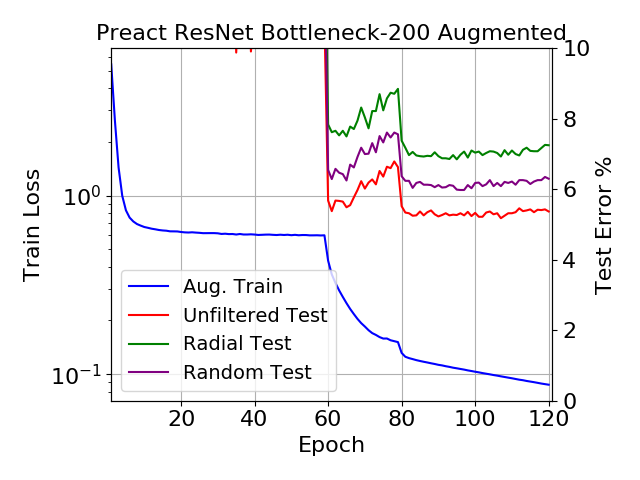}
  \caption[]%
  {{\small Preact-ResNet-200 trained on Fully Augmented CIFAR-10. Best viewed in color.}}    
  \label{fig:resnet-200-cifar10-augmented}
\end{figure}
    
\begin{figure}
  \captionsetup[subfigure]{justification=centering}
  \centering
  \begin{subfigure}[b]{\linewidth}
    \centering
    % stuff
    \begin{tabular}{|c|c|c|c|c|}
      \hline
      Train/Test & Unfilt. & Radial & Random & Gen. Gap\\ \hline
      Unfilt. & 5.54\% & 25.75\% & 12.31\% & 20.21\%\\ \hline
      Radial & 6.91\% & 7.91\% & 18.45\% & 11.54\% \\ \hline
      Random & 7.12\% & 35.03\% & 6.76\% & 28.27\%\\ \hline
      Augmented & 5.85\% & 7.89\% & 6.74\% & 2.04\% \\
      \hline
    \end{tabular}
    \caption[]%
    {{\small Preact-ResNet-Bottleneck-92 CIFAR-10 Generalization}}    
    \label{fig:resnet-92-cifar-10-table}
  \end{subfigure}
  \hfill
  \begin{subfigure}[b]{\linewidth}
    \centering
    % stuff
    \begin{tabular}{|c|c|c|c|c|}
      \hline
      Train/Test & Unfilt. & Radial & Random & Gen. Gap\\ \hline
      Unfilt. & 5.22\% & 23.37\% & 11.26\% &  18.15\%\\ \hline
      Radial & 6.35\% & 7.07\% & 17.09\% & 10.74\%\\ \hline
      Random & 6.47\% & 34.19\% & 5.9\% & 28.29\%\\ \hline
      Augmented & 5.37\% & 7.25\% & 6.3\% & 1.88\% \\
      \hline
    \end{tabular}
    \caption[]%
    {{\small Preact-ResNet-Bottleneck-200 CIFAR-10 Generalization}}   
    \label{fig:resnet-200-cifar-10-table}
  \end{subfigure}
  \caption[]%
  {\small CIFAR-10 Generalization Table. The train indicates what the training set was, the test indicates the test set. Final test 
  error is listed and the corresponding generalization gap.} 
  \label{fig:cifar-10-generalization-table}
\end{figure}

The CIFAR-10 generalization experiments otherwise are exactly the same setup as the SVHN generalization experiments from the previous 
section. In Figure \ref{fig:resnet-200-cifar-10} we present the generalization plots for the Preact-ResNet-200 trained on the unfiltered, 
randomly filtered and radially filtered training sets. In Figure \ref{fig:resnet-200-cifar10-augmented} we present the generalization plots 
for the Preact-ResNet-200 trained on the fully augmented CIFAR-10 dataset, and in Figure \ref{fig:cifar-10-generalization-table} we 
summarize all the exact error rates for the CIFAR-10 experiments. 

From these figures we observe that when trained on the unfiltered data, the networks exhibited a generalization gap when tested on the 
radially filtered test set, of about 18-20\%, much larger than the analogous gap for the SVHN dataset. Furthermore, when the nets were 
trained on the randomly filtered data, these networks again had the worst generalization gap at over 28\%, again for the 
radially filtered test set. The networks trained on the radially filtered data tend to 
have the lowest generalization gap, and furthermore that training on the augmented training set reduces the generalization gap. 
Similar to the SVHN experimental results, depth seemed to have a negligible effect on closing the generalization gap. 
    
\subsection{Discussion}

We now wish to synthesize the experimental results presented in the previous sections. First we extend our claim that human object 
recognizability is robust to Fourier filtered image data to the claim that: the neural networks trained on the Fourier filtered datasets 
(both the radial and random filtered datasets) actually generalized quite well to the unfiltered test set. Indeed, from the tables in 
Figures \ref{fig:svhn-generalization-table} and \ref{fig:cifar-10-generalization-table}, we highlight the fact that the nets that were 
trained on the random and radial datasets were only off by 1-2\% of the best unfiltered test accuracy. This suggests that our choice of 
Fourier filtering schemes produced datasets that are perceptually not too far off from the original unfiltered dataset. 

Despite the differences of the Fourier image statistics of the SVHN and CIFAR-10 datasets, as we noted previously, our SVHN and 
CIFAR-10 generalization experiments were of a nearly identical qualitative nature. We see that deep CNNs trained on an unfiltered natural 
image dataset exhibit a tendency to latch onto the image statistics of the training set, yielding a non-trivial generalization gap. The 
degree of this generalization gap can vary, ranging from 7-8\% for the SVHN to over 18\% for the CIFAR-10. Depth does not seem to have  
any real effect on reducing the observed generalization gaps. More generally, we note that there is no particular training set which 
generalizes universally well to all the test sets, though the radially filtered train set did tend to have the smaller generalization gap.

When training on the fully augmented training set, we observe an improvement in the generalization gap. However, we cast doubt on the 
notion that this sort of data augmentation scheme is sufficient to learn higher level semantic features in the dataset. Rather it is far 
more likely that the CNNs are learning a superficial robustness to the varying image statistics. To this end, we draw an analogy to 
adversarial training: augmenting the training set with a specific subset of adversarial examples does not make the network immune to 
adversarial examples in general. 

\section{Conclusion}

We are motivated by an oddity of CNNs: on the one hand they exhibit excellent generalization performance on difficult visual tasks, while 
on the other hand they exhibit an extreme sensitivity to adversarial examples. This sensitivity to adversarial examples suggests that these 
CNNs are not learning semantic concepts in the dataset. The goal of this article is to understand how a machine learning model can 
manage to generalize well without actually learning any high level semantics. 

Drawing upon computer vision literature on the statistical regularity of natural images, we believe that it is possible for natural image 
training and test datasets to share many superficial cues. By learning these superficial cues, a machine learning model would be able to 
sidestep the issue of high level concept learning and generalize well. To this end, we posed our main hypothesis: \emph{the current 
incarnation of deep neural networks have a tendency to learn surface statistical regularities as opposed to high level abstractions}. 

To measure this tendency, we claim it is sufficient to construct a map $F$ that perturbs a dataset in such a way that: 1) the 
recognizability of the objects/high level abstractions are almost entirely preserved from a human perspective while 2) the clean and 
perturbed datasets differ only in terms of their superficial statistical regularities. In this article, we show that appropriately tuned 
Fourier filtering satisfies these properties. 

In our experimental results, we show that CNNs trained on a dataset with one class of Fourier image statistics in general do not 
generalize universally well to test distributions exhibiting qualitatively different types of Fourier image statistics. In some cases we 
are able to show an up to 28\% gap in test accuracy. Furthermore, increasing the depth does not have a significant effect on closing this 
so-called generalization gap. We believe that this provides evidence for our main hypothesis.

While training on the fully augmented training set with the unfiltered and Fourier filtered datasets does have a significant impact on 
closing the generalization gap, we do not believe that this sort of data augmentation is sufficient for learning higher level abstractions 
in the dataset. It may be possible to generate some other perturbation of the dataset that yields a new generalization gap. 

With respect to promising new directions to solve the high level abstraction learning problem, recent work like \cite{ICF, UnsupAux} 
aim to learn good disentangled feature representations by combining unsupervised learning and reinforcement learning. In \cite{SCAN} a 
variational setup is used to learn visual concepts and \cite{DisentScene} aims to learn abstract relations between objects in natural scene 
images. More generally, new proposals such as \cite{ConsciousnessPrior} aim to transition away from making predictions in the perceptual 
space and instead operate in the higher order abstract space. We believe these are all novel directions towards a deep neural architecture 
that can learn high level abstractions. 

% I will remove this before the submission, leaving it in to calculate the page limits. 

\section*{Acknowledgements} 
We would like to acknowledge the developers of Theano \cite{Theano}. We would like to acknowledge the following 
organizations for their generous research funding and/or computational support (in alphabetical order): the CIFAR, Calcul Qu\'{e}bec, 
Canada Research Chairs, Compute Canada, the IVADO and the NSERC.

{\small
\bibliographystyle{ieee}
\bibliography{egbib}
}

\begin{appendices}
 
\section{Additional Fourier Filtered Images}

\subsection{SVHN}
\begin{figure}[ht!]
  \centering
  \includegraphics[width=0.95\linewidth]{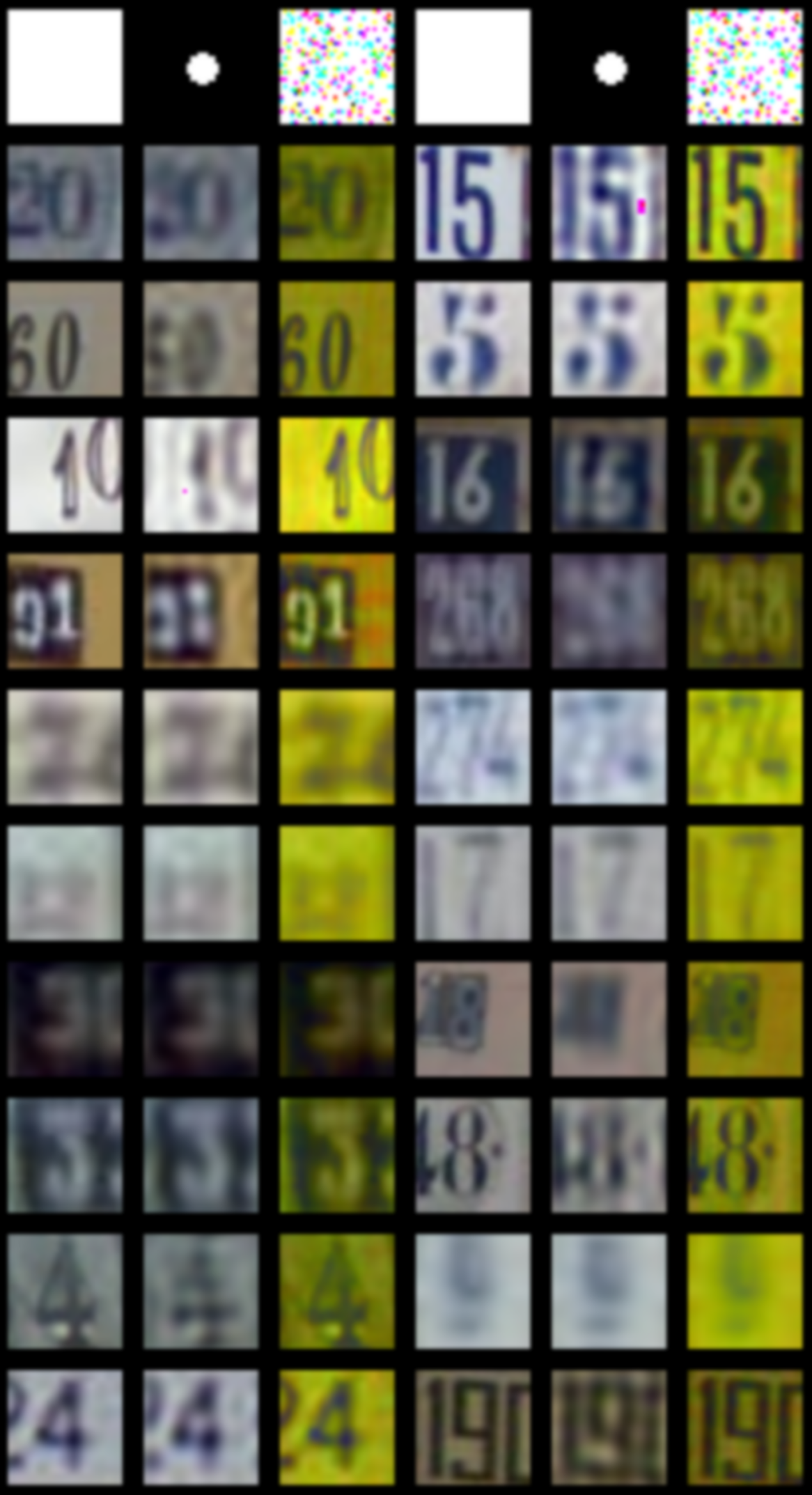}
  \caption[]%
  {{\small The first image in each row corresponds to the Fourier mask in frequency space. A white pixel corresponds to preserving the 
Fourier 
mode, black/color corresponds to setting it to zero. Best viewed in color.}}    
  \label{fig:additional-svhn}
\end{figure}

In Figure \ref{fig:additional-svhn} we present more pictures of Fourier filtered SVHN images. We present a random mask which can 
result in color deformations. We show 2 randomly chosen images for each label. 
% Long SVHN strip:

\subsection{CIFAR-10} 

\begin{figure}[ht!]
  \centering
  \includegraphics[width=0.95\linewidth]{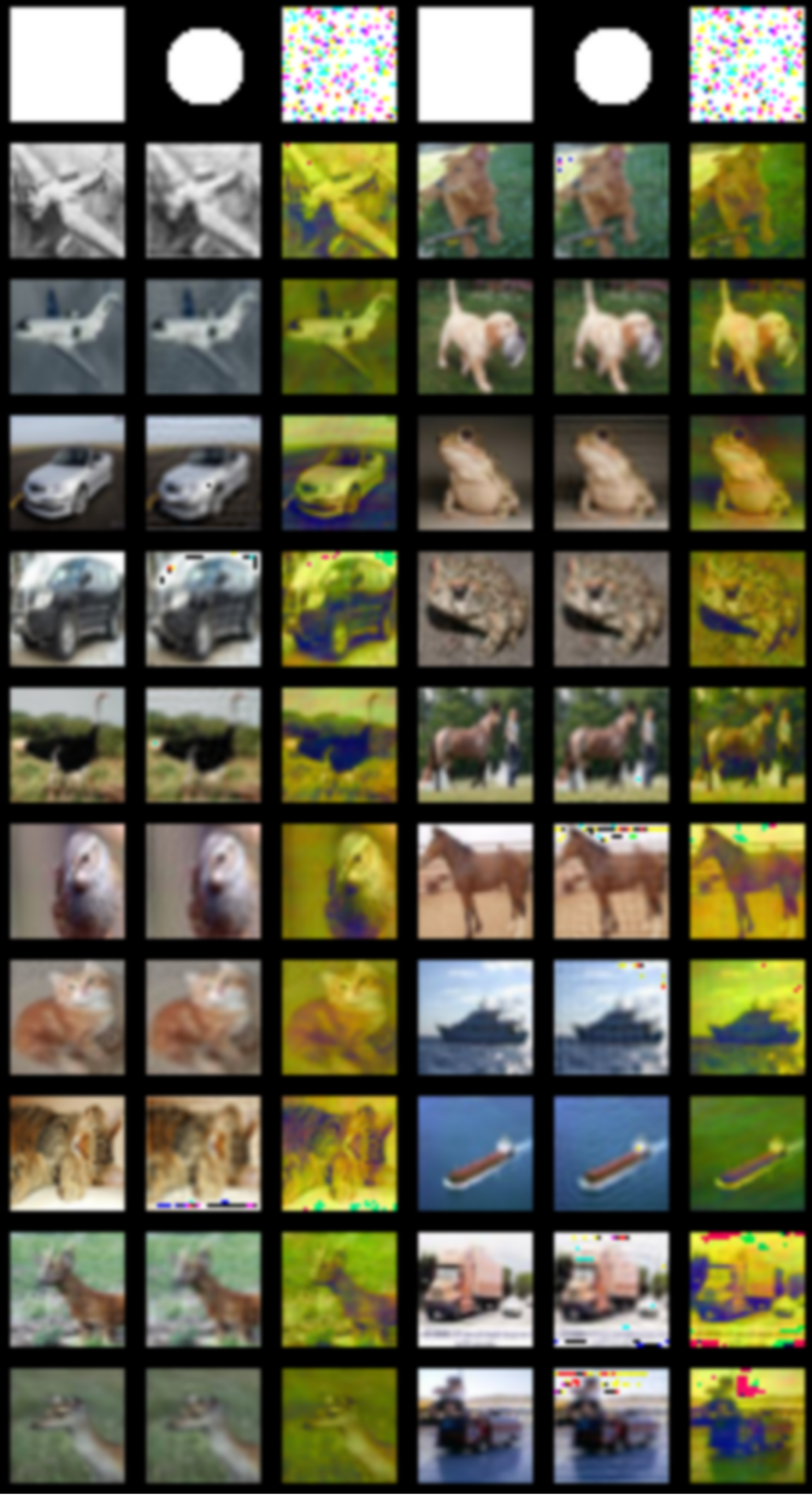}
  \caption[]%
  {{\small The first image in each row corresponds to 
a Fourier mask in frequency space. A white pixel corresponds to preserving the 
Fourier 
mode, black/color corresponds to setting it to zero. Best viewed in color.}}    
  \label{fig:additional-cifar10}
\end{figure}

In Figure \ref{fig:additional-cifar10} we present more pictures of Fourier filtered CIFAR-10 images. We present a random mask which can 
result in color deformations. We show 2 randomly chosen images for each label. 

% Long CIFAR-10 strip:

\section{Preact-ResNet-92 Experimental Plots}

In this section we share our Preact-ResNet-92 graphical plots.

\subsection{SVHN} 

\begin{figure}
  \centering
  \begin{subfigure}[b]{0.95\linewidth}
    \centering
    \includegraphics[width=\linewidth]{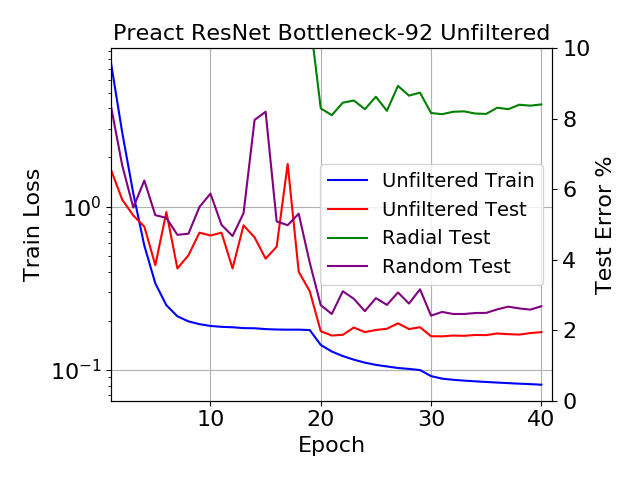}
    \caption[]%
    {{\small Trained on Unfiltered SVHN}}    
    \label{fig:resnet-92-svhn-unfiltered}
  \end{subfigure}
  \hfill
  \begin{subfigure}[b]{0.95\linewidth}
    \centering
    \includegraphics[width=\linewidth]{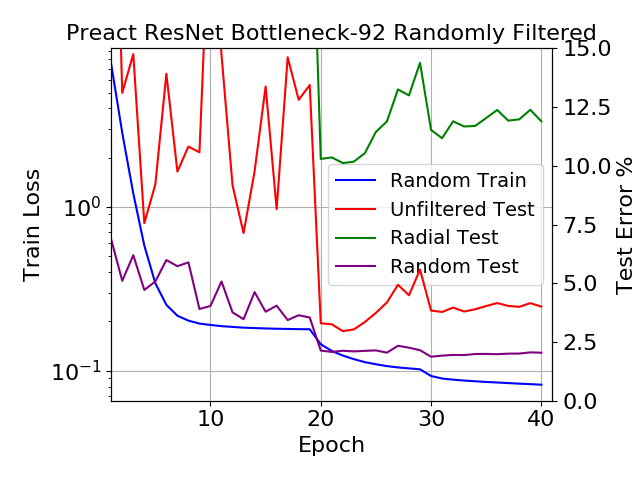}
    \caption[]%
    {{\small Trained on Randomly Filtered SVHN}}    
    \label{fig:resnet-92-svhn-random}
  \end{subfigure}
  \hfill
  \begin{subfigure}[b]{0.95\linewidth}   
  \centering 
    \includegraphics[width=\linewidth]{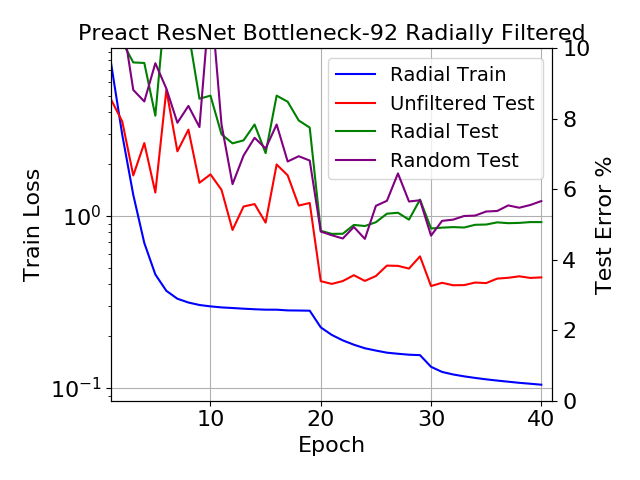}
    \caption[]%
    {{\small Trained on Radially Filtered SVHN}}    
    \label{fig:resnet-92-svhn-radial}
  \end{subfigure}
  \caption[]%
  {\small Generalization plots for Preact-ResNet-Bottleneck-92 model. \textbf{(a)} Trained on unfiltered SVHN. \textbf{(b)} 
  Trained on randomly filtered SVHN data. \textbf{(c)} Trained on radially filtered SVHN data. Best viewed in color.} 
  \label{fig:resnet-92-svhn}
\end{figure}

\begin{figure}[ht!]
  \centering
  \includegraphics[width=0.95\linewidth]{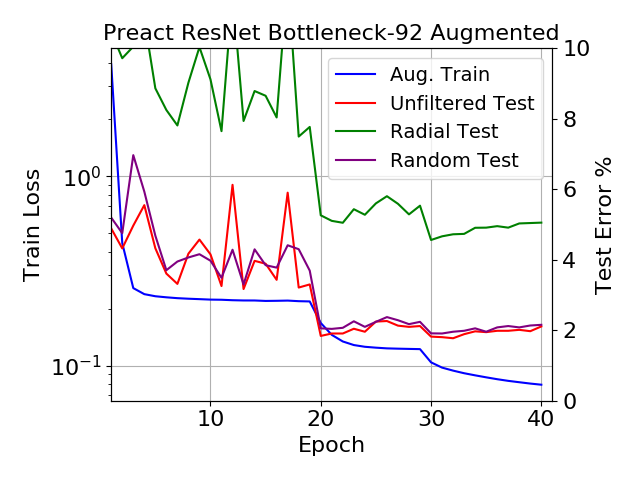}
  \caption[]%
  {{\small Preact-ResNet-92 Trained on Fully Augmented SVHN. Best reviewed in color.}}    
  \label{fig:resnet-92-svhn-augmented}
\end{figure}

In Figures \ref{fig:resnet-92-svhn} and \ref{fig:resnet-92-svhn-augmented} we show the Preact-ResNet-92 plots for the SVHN datasets.

% ResNet-92 SVHN 

% ResNet-92 SVHN fully augmented:

\subsection{CIFAR-10} 

\begin{figure}[ht!]
  \centering
  \begin{subfigure}[b]{0.95\linewidth}
    \centering
    \includegraphics[width=\linewidth]{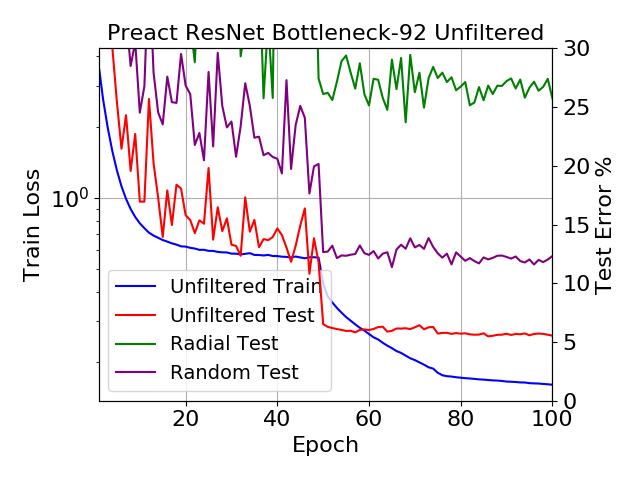}
    \caption[]%
    {{\small Trained on Unfiltered CIFAR-10}}    
    \label{fig:resnet-92-cifar-10-unfiltered}
  \end{subfigure}
  \hfill
  \begin{subfigure}[b]{0.95\linewidth}
    \centering
    \includegraphics[width=\linewidth]{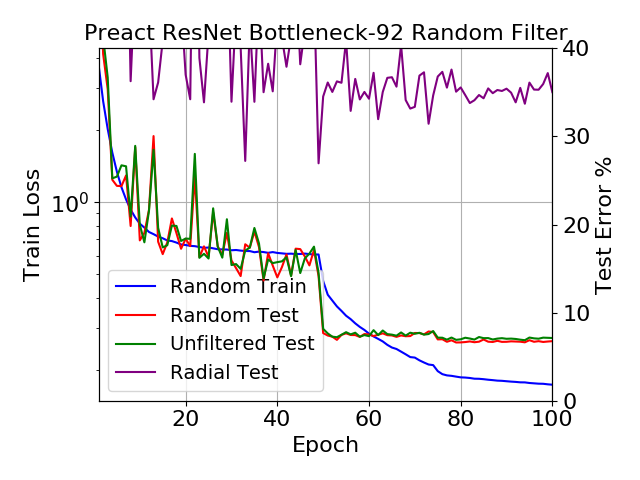}
    \caption[]%
    {{\small Trained on Randomly Filtered CIFAR-10}}    
    \label{fig:resnet-92-cifar-10-random}
  \end{subfigure}
  \hfill
  \begin{subfigure}[b]{0.95\linewidth}   
    \centering 
    \includegraphics[width=\linewidth]{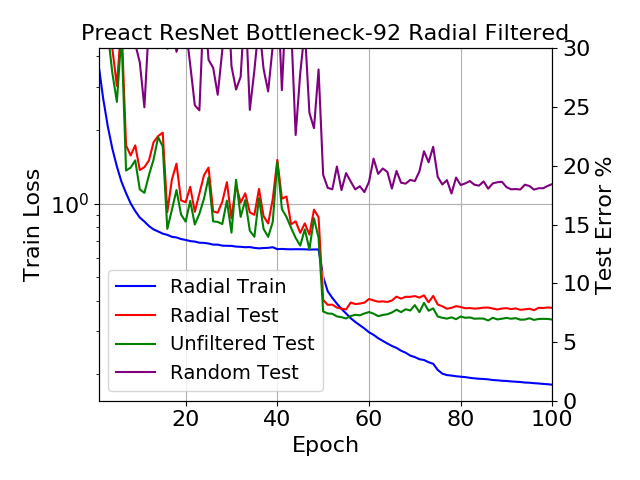}
    \caption[]%
    {{\small Trained on Radially Filtered CIFAR-10}}    
    \label{fig:resnet-92-cifar-10-radial}
  \end{subfigure}
  \caption[]%
  {\small Generalization plots for Preact-ResNet-Bottleneck-92 model. \textbf{(a)} Trained on unfiltered CIFAR-10 data. 
  \textbf{(b)} Trained on randomly filtered CIFAR-10 data. \textbf{(c)} Trained on radially filtered CIFAR-10 data. Best viewed in color.} 
  \label{fig:resnet-92-cifar-10}
\end{figure}

\begin{figure}[ht!]
  \centering
  \includegraphics[width=0.95\linewidth]{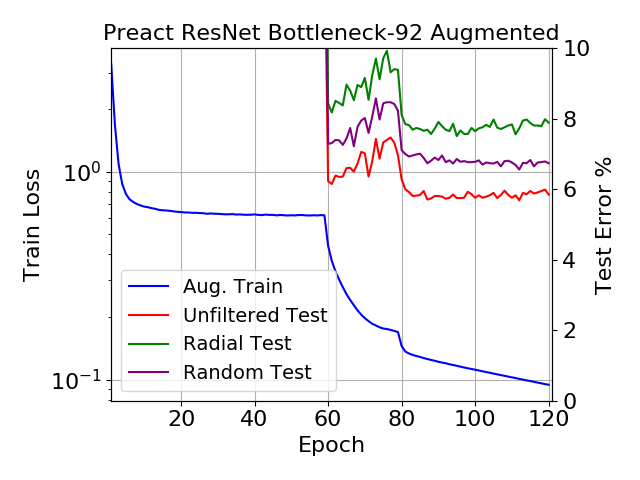}
  \caption[]%
  {{\small Preact-ResNet-92 Trained on Fully Augmented CIFAR-10. Best viewed in color.}}    
  \label{fig:resnet-92-cifar10-augmented}
\end{figure}

In Figures \ref{fig:resnet-92-cifar-10} and \ref{fig:resnet-92-cifar10-augmented} we show the Preact-ResNet-92 plots for the SVHN 
datasets.

% ResNet-92 CIFAR-10:
    
% ResNet-92 Fully Augmented CIFAR-10:

\end{appendices}

\end{document}